# EpiScreen: Early Epilepsy Detection from Electronic Health Records with Large Language Models

Shuang Zhou[1], Kai Yu[1], Zaifu Zhan[2], Huixue Zhou[1], Min Zeng[1], Feng Xie[1], Zhiyi Sha[3], Rui Zhang[1],*

**Affiliations:**
1. Division of Computational Health Sciences, Department of Surgery, University of Minnesota, Minneapolis, MN, USA
2. College of Science and Engineering, University of Minnesota, Minneapolis, MN, USA
3. Department of Neurology, University of Minnesota, Minneapolis, MN, USA

*Correspondence: ruizhang@umn.edu

# Abstract

Epilepsy and psychogenic non-epileptic seizures often present with similar seizure-like manifestations but require fundamentally different management strategies. Misdiagnosis is common and can lead to prolonged diagnostic delays, unnecessary treatments, and substantial patient morbidity. Although prolonged video-electroencephalography is the diagnostic gold standard, its high cost and limited accessibility hinder timely diagnosis. Here, we developed a low-cost, effective approach, EpiScreen, for early epilepsy detection by utilizing routinely collected clinical notes from electronic health records. Through fine-tuning large language models on labeled notes, EpiScreen achieved an AUC of up to 0.875 on the MIMIC-IV dataset and 0.980 on a private cohort of the University of Minnesota. In a clinician-AI collaboration setting, EpiScreen-assisted neurologists outperformed unaided experts by up to 10.9%. Overall, this study demonstrates that EpiScreen supports early epilepsy detection, facilitating timely and cost-effective screening that may reduce diagnostic delays and avoid unnecessary interventions, particularly in resource-limited regions.

# Introduction

Epilepsy is among the most prevalent chronic neurological disorders, affecting approximately 50 million people worldwide, including about 2.9 million individuals with active epilepsy in the United States [1,2]. Psychogenic non-epileptic seizures (PNES) and epilepsy are clinically distinct conditions, yet they often present with similar paroxysmal events, creating substantial diagnostic uncertainty [3,4]. Their underlying pathophysiology, treatment strategies, and prognoses differ fundamentally, making accurate and timely differentiation critical for effective management [5,6]. Nevertheless, misdiagnosis remains common; prior studies indicate that 20–30% of patients referred to tertiary epilepsy centers for presumed drug-resistant epilepsy are ultimately diagnosed with PNES [7,8].

In routine clinical practice, patients with seizure-like events typically first present to primary care and are subsequently referred to specialty care (e.g., neurology) when epilepsy is suspected. During specialist evaluation, neurologists integrate multiple sources of clinical information, including detailed history-taking, witness descriptions, neurological and general physical examinations, laboratory testing, routine scalp electroencephalography (EEG), and neuroimaging such as magnetic resonance imaging (MRI) [9,10]. While these assessments provide a comprehensive clinical profile to support diagnostic decision-making, prolonged video-electroencephalography (vEEG) monitoring remains the diagnostic gold standard because it enables simultaneous correlation of clinical events with electrographic activity [11]. However, vEEG is costly, resource-intensive, and logistically demanding. In the United States, a single vEEG admission commonly exceeds $15,000, creating a substantial financial burden for patients and healthcare systems [12]. Moreover, vEEG is typically performed in specialized inpatient epilepsy monitoring units with limited capacity, leading to access disparities. Typical wait times of 8–12 weeks, combined with monitoring durations of 1–5 days, can extend the diagnostic process over a few months, potentially delaying a definitive diagnosis and appropriate treatment [13,14].

Given the cost and burden of vEEG, there is strong motivation to explore alternative diagnostic approaches that are both accessible and efficient [15–17]. Clinical notes, routinely documented in electronic health records (EHRs), represent a rich and feasible data source [18]. These narratives capture detailed patient information, including seizure semiology, medical history, physical examination findings, and laboratory results. Leveraging the latent diagnostic signals in these texts may enable earlier and accurate differentiation between epilepsy and PNES, thus reducing diagnostic delays, avoiding unnecessary treatments, and improving patient outcomes [19–21].

Although several studies have explored leveraging clinical notes for epilepsy detection, their diagnostic performance has remained limited [22,23]. One major challenge arises from the probabilistic nature of seizure-related clinical features. While experts have identified a range of semiologic and historical features, such as ictal duration, eye closure, or tongue biting, that are statistically more common in epilepsy or PNES, none of these features is pathognomonic [9,24]. Many PNES patients display signs typically associated with epileptic seizures, and conversely, some epilepsy patients may lack classical epileptic markers [25]. Consequently, rule-based or feature-dependent models, as well as general-purpose large language models (LLMs), often yield suboptimal performance [16,26,27]. An additional challenge stems from the intrinsic subjectivity of clinical narratives. Descriptions of seizures are frequently derived from patients, family members, or eyewitnesses, whose observations can be incomplete or imprecise. Therefore, terms such as "loss of consciousness", "generalized shaking", or "unresponsiveness" may appear in both conditions, obscuring diagnostic boundaries [28,29]. These ambiguities underscore the importance of analyzing the entire clinical narrative in a holistic and context-aware manner [30].

To address these challenges, we developed a customized artificial intelligence (AI) model, EpiScreen, for early epilepsy detection using clinical narratives (Fig. 1). Through fine-tuning LLMs on labeled patient narratives, EpiScreen learned nuanced linguistic cues, contextual relationships, and latent semantic distinctions to detect epilepsy. Results showed that EpiScreen achieved an AUC of up to 0.875 (95% CI: 0.848–0.899) and 0.980 (95% CI: 0.974–0.985) on the MIMIC-IV dataset [31] and a private cohort from the University of Minnesota, while its cross-institutional performance surpassed the established comparative method by over 14.2% and 24.2%, respectively. Furthermore, in clinician-AI collaboration settings, EpiScreen-assisted clinicians outperformed independent clinicians up to 10.9%. Overall, our study demonstrates the effectiveness of leveraging routinely collected clinical notes for epilepsy detection, providing a low-cost, scalable, and efficient approach that could facilitate earlier diagnosis and improve clinical decision-making.

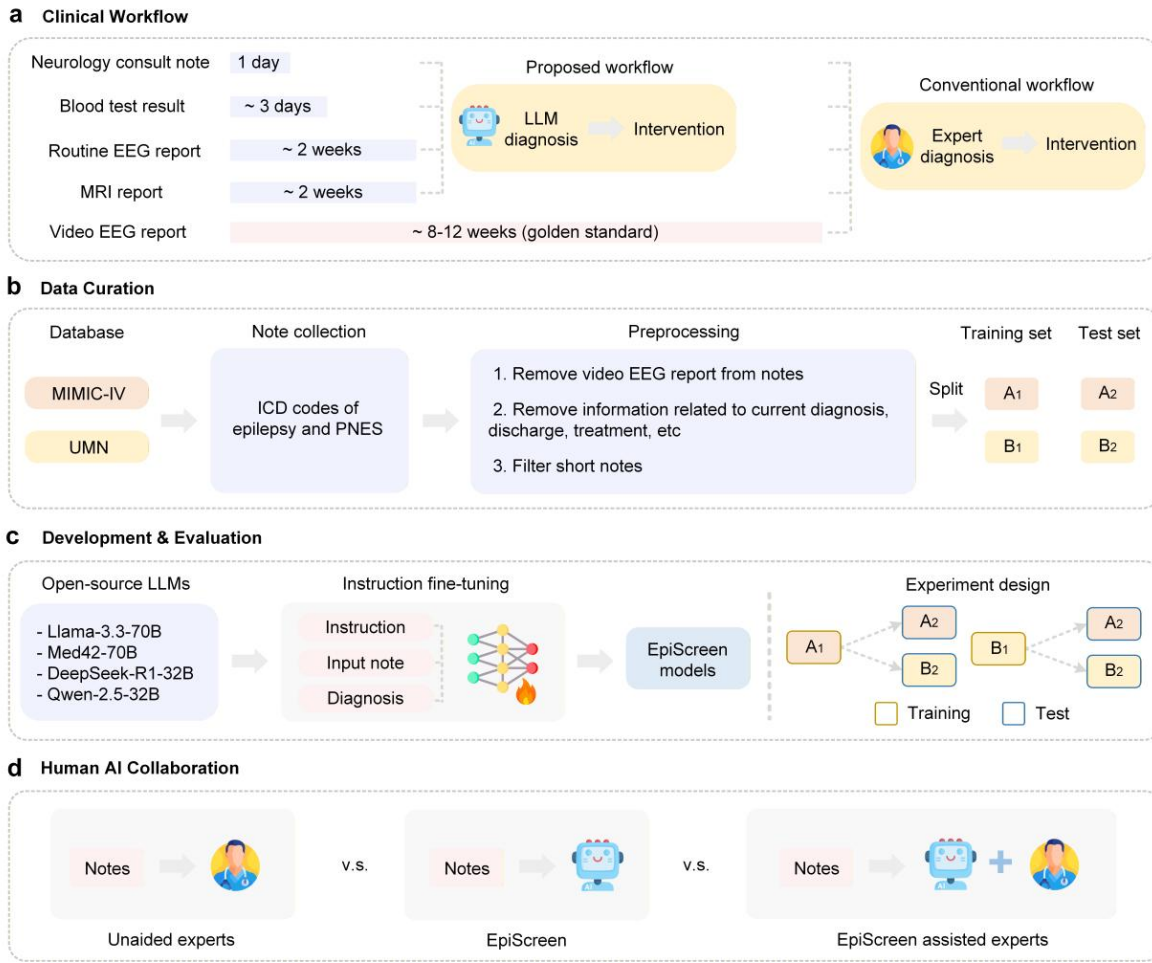

**Fig. 1 | Overview of the EpiScreen framework. a,** Comparison between the conventional epilepsy diagnosis workflow and the proposed EpiScreen-enabled workflow. In routine practice, long-term video EEG monitoring, the gold standard for epilepsy diagnosis, is resource-intensive and often requires 8–12 weeks for scheduling and reporting after referral to neurology. EpiScreen is designed to operate at this specialty-care stage, leveraging routinely documented clinical narratives and reports generated during and after neurology referral to provide earlier diagnostic support prior to confirmatory video EEG. Timeline estimates are based on data from the University of Minnesota (UMN) hospitals and may vary across healthcare systems. **b,** Data curation pipeline. Two electronic health record datasets were used: the publicly available MIMIC-IV dataset and an independent private cohort from UMN. **c,** Model development and evaluation framework. EpiScreen was trained and validated using a cross-institutional design to assess robustness and generalizability across healthcare settings. **d,** Human–AI collaboration analysis. Diagnostic performance was compared among unaided clinical experts, EpiScreen alone, and experts assisted by EpiScreen, demonstrating the added value of integrating the model into clinical practice.

# Results

## Datasets

We evaluated model performance using EHRs from the MIMIC-IV database [31] and the University of Minnesota Clinical Data Repository (UMN). Dataset statistics and composition are summarized in Table 1 and Supplementary Tables S1–2.

**Table 1. Data statistics of the MIMIC-IV and UMN datasets.**

| Dataset | MIMIC-IV | UMN |
|---|---|---|
| Sample size (overall, epilepsy, PNES), n | (3451, 2651, 800) | (10182, 8011, 2171) |
| Age (y), mean ± s.d. | 52.0 ± 19.3 | 49.6 ± 21.1 |
| Male (%) | 53.1% | 51.0% |
| Race (White, Black, Asian, others), % | (65.6%, 15.9%, 1.9%, 16.6%) | (75.4%, 12.1%, 3.0%, 9.4%) |
| Overall note length (words), mean ± s.d. | 1199.0 ± 270.5 | 970.0 ± 297.3 |
| Epilepsy note length (words), mean ± s.d. | 1191.1 ± 270.6 | 950.4 ± 293.8 |
| PNES note length (words), mean ± s.d. | 1224.7 ± 268.7 | 1045.1 ± 299.0 |

**Table 2. Epilepsy detection performance comparison.** Model training and testing were conducted using data from the same institution. Values are reported along with 95% confidence interval.

| Model | MIMIC-IV | | UMN | |
|---|---|---|---|---|
| | AUC | Accuracy | AUC | Accuracy |
| FSLS | 0.695 [0.648, 0.736] | 0.656 [0.619, 0.692] | 0.944 [0.934, 0.955] | 0.790 [0.773, 0.808] |
| BERT | 0.756 [0.715, 0.797] | 0.773 [0.740, 0.805] | 0.952 [0.941, 0.963] | 0.919 [0.906, 0.931] |
| ClinicalBERT | 0.769 [0.732, 0.808] | 0.760 [0.728, 0.792] | 0.960 [0.950, 0.970] | 0.889 [0.875, 0.903] |
| Qwen-2.5-32B (prompt) | 0.789 [0.752, 0.825] | 0.745 [0.711, 0.777] | 0.900 [0.881, 0.918] | 0.882 [0.867, 0.896] |
| DeepSeek-R1-32B (prompt) | 0.748 [0.704, 0.791] | 0.768 [0.738, 0.802] | 0.848 [0.827, 0.867] | 0.787 [0.769, 0.805] |
| Llama-3.3-70B (prompt) | 0.813 [0.779, 0.846] | 0.645 [0.611, 0.680] | 0.894 [0.875, 0.911] | 0.867 [0.852, 0.882] |
| Med42-70B (prompt) | 0.787 [0.749, 0.824] | 0.786 [0.755, 0.816] | 0.885 [0.871, 0.899] | 0.836 [0.820, 0.850] |
| EpiScreen (Qwen-2.5-32B) | 0.887 [0.862, 0.911] | **0.835 [0.807, 0.864]** | **0.992 [0.989, 0.995]** | **0.967 [0.958, 0.974]** |
| EpiScreen (DeepSeek-R1-32B) | 0.882 [0.856, 0.907] | 0.825 [0.800, 0.852] | 0.992 [0.988, 0.995] | 0.961 [0.952, 0.969] |
| EpiScreen (Llama-3.3-70B) | 0.889 [0.863, 0.913] | 0.831 [0.803, 0.857] | 0.974 [0.966, 0.981] | 0.936 [0.924, 0.946] |
| EpiScreen (Med42-70B) | **0.891 [0.866, 0.915]** | 0.832 [0.805, 0.860] | 0.976 [0.967, 0.985] | 0.965 [0.957, 0.973] |

**Table 3. Generalization performance of epilepsy detection models across institutions.** Training and test data were obtained from different institutions. Models trained on the UMN dataset were evaluated on the MIMIC-IV dataset (left), whereas models trained on the MIMIC-IV dataset were evaluated on the UMN dataset (right). Values are reported along with 95% confidence interval.

| Model | MIMIC-IV | | UMN | |
|---|---|---|---|---|
| | AUC | Accuracy | AUC | Accuracy |
| FSLS | 0.498 [0.444, 0.547] | 0.232 [0.198, 0.260] | 0.485 [0.456, 0.516] | 0.568 [0.545, 0.590] |
| BERT | 0.699 [0.652, 0.743] | 0.504 [0.469, 0.541] | 0.701 [0.675, 0.729] | 0.760 [0.741, 0.777] |
| ClinicalBERT | 0.749 [0.709, 0.786] | 0.616 [0.578, 0.654] | 0.788 [0.763, 0.811] | 0.806 [0.788, 0.823] |
| EpiScreen (Qwen-2.5-32B) | 0.867 [0.842, 0.891] | 0.740 [0.706, 0.771] | 0.979 [0.974, 0.984] | **0.918 [0.907, 0.930]** |
| EpiScreen (DeepSeek-R1-32B) | **0.875 [0.848, 0.899]** | 0.700 [0.667, 0.734] | 0.966 [0.959, 0.974] | 0.910 [0.898, 0.922] |
| EpiScreen (Llama-3.3-70B) | 0.829 [0.798, 0.858] | **0.770 [0.741, 0.803]** | 0.978 [0.973, 0.984] | 0.894 [0.882, 0.907] |
| EpiScreen (Med42-70B) | 0.850 [0.821, 0.876] | 0.734 [0.699, 0.767] | **0.980 [0.974, 0.985]** | 0.917 [0.906, 0.927] |

## Epilepsy detection performance

We evaluated the epilepsy detection performance of EpiScreen against established methods and off-the-shelf LLMs (Table 2). On the MIMIC-IV dataset, the phenotype-based FSLS method [19] achieved an AUC of 0.695 (95% CI: 0.648–0.736), while conventional language models such as ClinicalBERT [32] achieved a higher AUC of 0.769 (95% CI: 0.732–0.808). A similar pattern was observed on the UMN dataset, where ClinicalBERT achieved an AUC of 0.960 (95% CI: 0.950–0.970), exceeding the FSLS's AUC of 0.944 (95% CI: 0.934–0.955). In contrast, EpiScreen consistently outperformed all baseline approaches across datasets. The model fine-tuned on Med42-70B [33] achieved an AUC of 0.891 (95% CI: 0.866–0.915) on MIMIC-IV and 0.976 (95% CI: 0.967–0.985) on UMN. Importantly, fine-tuning substantially improved performance relative to off-the-shelf LLMs, yielding average absolute AUC gains of 13.2% on MIMIC-IV and 11.6% on UMN.

## Generalization performance

To evaluate cross-institutional generalizability [34], we assessed model performance on held-out external test data (Table 3). The comparative models exhibited marked performance degradation when transferred across institutions. Specifically, the phenotype-based FSLS model [19] trained on UMN data achieved an AUC of 0.498 (95% CI: 0.444–0.547) on MIMIC-IV, representing a 47.2% relative decline compared to its in-domain performance. Similarly, the BERT model [30] trained on MIMIC-IV achieved an AUC of 0.701 (95% CI: 0.675–0.729) on UMN, corresponding to a 26.6% performance reduction relative to internal validation. In contrast, EpiScreen demonstrated substantially improved cross-site robustness. When trained on UMN data using Qwen-2.5-32B [35], EpiScreen achieved an AUC of 0.867 (95% CI: 0.842–0.891) on MIMIC-IV, with only a 12.6% relative decrease compared to its in-domain performance. Conversely, EpiScreen models trained on MIMIC-IV generalized strongly to UMN, achieving an average AUC of 0.979, substantially exceeding the cross-institutional performance of ClinicalBERT (AUC 0.788, 95% CI: 0.763–0.811) and BERT (AUC 0.701, 95% CI: 0.675–0.729). Overall, EpiScreen improved cross-institutional AUC by 14.2% and 24.2% relative to the established comparative models on the two external evaluations, respectively.

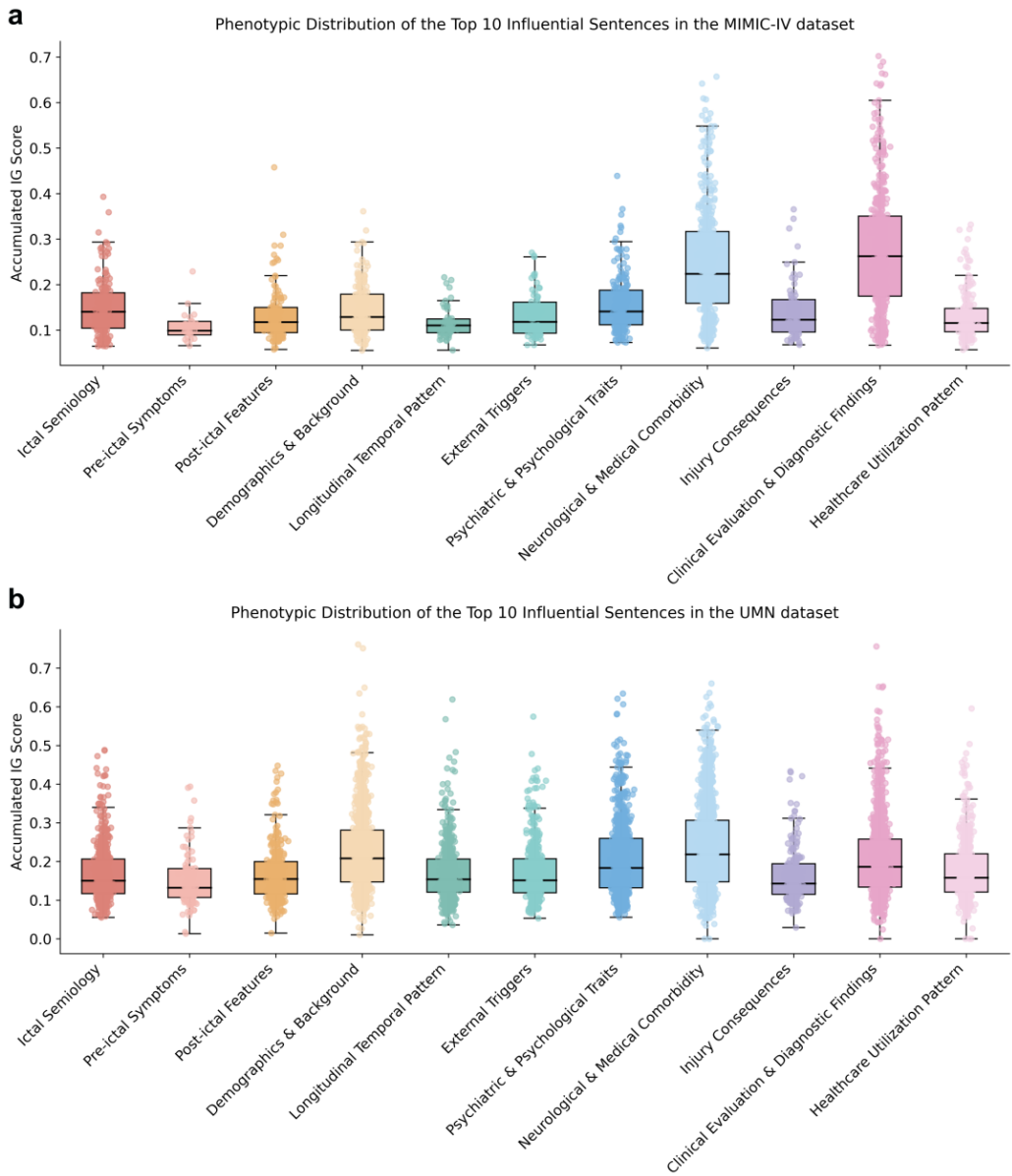

**Fig. 2 | Attribution scores for epilepsy identification.** Explainability results are shown for EpiScreen trained on Qwen-2.5-32B. Box plots show the distribution of Integrated Gradients scores for the top 10 most influential sentences across 11 phenotype categories, as derived from models trained on the MIMIC-IV (a) and UMN datasets (b). Accumulated IG score represents the normalized sum of integrated gradient attribution scores within a given phenotype category among the top-ranked sentences in each note, reflecting the relative contribution of that category to the model's prediction. Box plots show the median (central line), interquartile range (IQR; box), and whiskers extending to 1.5 × IQR. Individual data points are overlaid.

## Model explainability

We assessed model explainability using a sentence-level Integrated Gradients (IG) approach [36] to identify which components of clinical notes most strongly influenced epilepsy detection. The explainability results for EpiScreen trained on Qwen-2.5-32B are shown in Fig. 2 and Supplementary Figs. 1–2. In the MIMIC-IV cohort, the phenotype categories Neurological & Medical Comorbidity (e.g., migraine, stroke) and Diagnostic Testing Profile (e.g., normal EEG) demonstrated the strongest influence among the top 10 most influential sentences, with a median accumulated IG score exceeding 0.2 (Fig. 2a). Similarly, analysis of the UMN cohort, based on the upper quartile (75th percentile) of the IG distribution, identified Demographics & Background (e.g., age at presentation, family history), Psychiatric & Psychological Phenotype (e.g., depression, anxiety disorder), Neurological & Medical Comorbidity, and Diagnostic Testing Profile as the most influential feature categories for epilepsy detection (Fig. 2b).

To provide qualitative insight, we further present representative case examples from the MIMIC-IV dataset illustrating the top 10 most influential sentences (Table 4 and Supplementary Tables S4–7). As shown in Table 4, seizure semiology is frequently ranked among the highest-attribution sentences. For example, the description "developed acute onset of left facial redness, left eye fluttering, and rhythmic movements of the left arm lasting 45 minutes" reflects prolonged, stereotyped, and lateralized motor activity, which are features commonly associated with epileptic seizures, and received an IG score of 0.946. Pre-event features also contributed to the diagnosis. The sentence "These headaches are often followed by her aura (awareness that she may have an event)" describes a temporally structured sequence of symptoms (headache followed by aura) that is clinically compatible with a seizure prodrome. This sentence received an IG score of 0.930, indicating a strong influence on the model's prediction.

**Table 4. A case study from the MIMIC-IV dataset demonstrating model explainability for epilepsy detection.**

| Index | Sentence | IG Score | Phenotype Category |
|---|---|---|---|
| 1 | In ___, she was witnessed at ___ to develop acute onset of left facial redness, left eye fluttering and chronic movements of the left arm that lasted 45 minutes. | 0.946 | Seizure Semiology |
| 2 | The second one was much more prolonged and started again with a behavioral arrest but evolved into shaking of the arms and the third event occurred out of sleep and consisted of 30 seconds of shaking of the arms followed by behavioral arrest. | 0.932 | Seizure Semiology |
| 3 | These headaches are often followed by her aura (awareness that she may have an event), but if the headache is treated, she does not have an event. | 0.930 | Pre-Event Features |
| 4 | The patient was noted to have an event in the triage area and several more while in the ED proper for a total of nearly 10 events today. | 0.894 | Event Temporal Pattern |
| 5 | The background activity was relatively slow suggestive of a mild encephalopathy with additional focal slowing in the left central area suggestive of subcortical dysfunction in this region. | 0.878 | Diagnostic Testing & Evaluation Phenotype |
| 6 | Chronic microvascular ischemic disease. | 0.851 | Medical Comorbidity Profile |
| 7 | EEG ___: This is an abnormal routine EEG due to the slow background suggestive of a moderate encephalopathy. | 0.831 | Diagnostic Testing & Evaluation Phenotype |
| 8 | Language is fluent with intact repetition and comprehension but diminishes prosody. | 0.818 | Medical Comorbidity Profile |
| 9 | She does not recall the events exactly, but she noted an aura and extension of both arms with subsequent loss of consciousness. | 0.795 | Seizure Semiology |
| 10 | Past Medical History: Started in ___ - Fell from a horse as a child and had head injury. | 0.783 | Demographics & Background |

Note: The table presents the ten sentences from a single patient note with the highest sentence-level Integrated Gradient (IG) attribution scores. Each sentence is categorized into one of 11 predefined clinical phenotypes to interpret the salient features for the model's prediction. Explainability results are shown for EpiScreen trained on Qwen-2.5-32B.

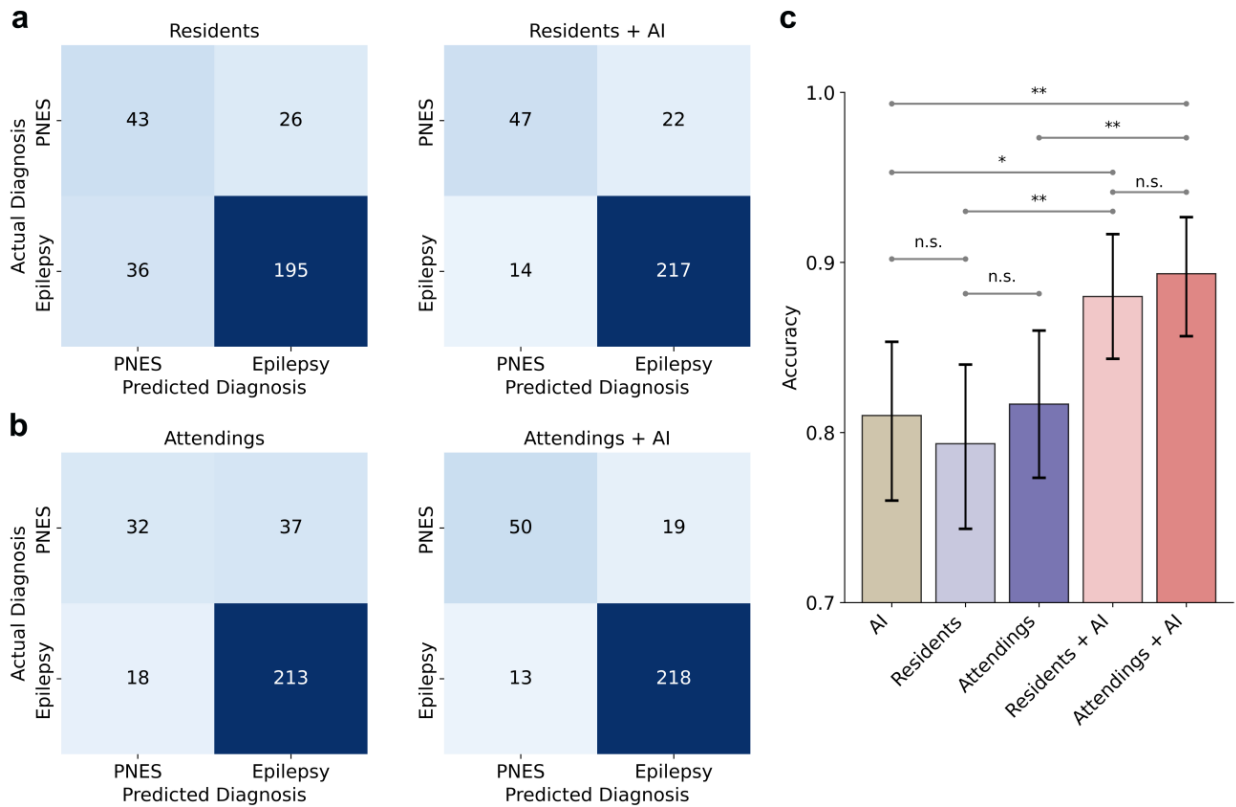

**Fig. 3 | Evaluation of clinician performance with AI support. a,** Confusion matrices comparing diagnostic decisions made by two resident physicians under consensus review versus the same residents assisted by AI. **b,** Confusion matrices comparing diagnostic decisions made by two attending physicians under consensus review versus the same attendings assisted by AI. **c,** Diagnostic accuracy for epilepsy classification across three conditions: EpiScreen alone (fine-tuned on Qwen-2.5-32B), clinicians without AI assistance, and clinicians supported by model-generated predictions and sentence-level Integrated Gradients attributions. Evaluations were conducted on a stratified subset of the MIMIC-IV test cohort (n = 300). Significance levels are indicated as \*\*\*$p < 0.001$, \*\*$p < 0.01$, \*$p < 0.05$, and n.s. ($p > 0.05$), based on two-sided Mann–Whitney U tests.

## AI-assisted clinician assessments

We next assessed whether EpiScreen could provide practical decision support for neurologists in epilepsy detection. A stratified random sample of 300 cases was drawn from the MIMIC-IV test cohort to preserve class distribution. Diagnostic performance was evaluated under three conditions (Fig. 1d): (i) EpiScreen operating independently (AI alone); (ii) clinicians working without assistance (residents alone and attendings alone); and (iii) clinicians supported by model outputs, including predicted labels and sentence-level Integrated Gradients attributions (AI-assisted residents and attendings).

As shown in Fig. 3c, the diagnostic accuracy of residents and attendings without AI support was 0.793 (95% CI: 0.743–0.840) and 0.817 (95% CI: 0.773–0.860), respectively, which did not differ significantly from the standalone EpiScreen performance (0.810, 95% CI: 0.760–0.853; $p > 0.05$). With AI assistance, accuracy improved substantially to 0.880 (95% CI: 0.843–0.917) for residents and 0.893 (95% CI: 0.857–0.927) for attendings. Error pattern analysis further demonstrated complementary benefits. AI assistance reduced false negatives among residents from 36 to 14 (Fig. 3a), while decreasing false positives among attendings from 37 to 19 (Fig. 3b). Overall, the human–AI collaborative approach achieved absolute accuracy gains of up to 10.9% compared with unaided clinicians ($p < 0.01$) and up to 10.2% relative to the standalone AI model ($p < 0.05$).

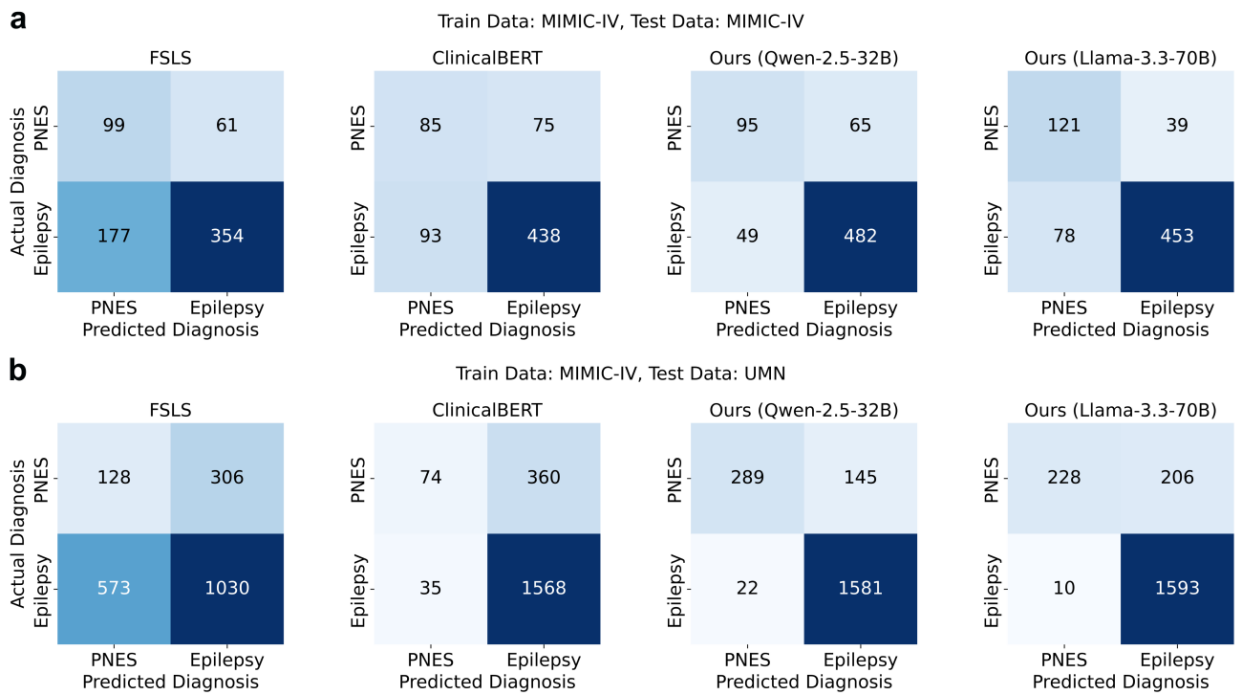

**Fig. 4 | Confusion matrix comparison for epilepsy identification. a,** Confusion matrices for epilepsy identification across models, with training and testing both conducted on MIMIC-IV. Rows denote actual diagnoses and columns denote predicted diagnoses. EpiScreen models yield fewer false negatives and more true positives than FSLS and ClinicalBERT, indicating higher sensitivity for epilepsy detection. **b,** Confusion matrices of models trained on MIMIC-IV and tested on the UMN dataset. EpiScreen models show fewer false negatives and false positives.

## Error analysis

We further examined confusion matrices to characterize model error patterns (Fig. 4 and Supplementary Fig. 3). For in-domain evaluation on the MIMIC-IV dataset, the comparative methods showed higher false-negative counts. Specifically, FSLS [19] and ClinicalBERT [30] produced 177 and 93 false negatives, respectively. In contrast, the fine-tuned Qwen-2.5-32B model produced 49 false negatives (Fig. 4a). For external validation on the UMN dataset, differences were more pronounced. The phenotype-based method FSLS yielded 573 false negatives, whereas the fine-tuned Llama-3.3-70B [37] model yielded 10. A similar trend was observed for false positives. On the UMN dataset, FSLS and ClinicalBERT produced 360 and 306 false positives, respectively, while EpiScreen showed lower counts, with 145 and 206 false positives for the two base models (Fig. 4b). Overall, EpiScreen achieved higher true-positive and true-negative counts while reducing both false-negative and false-positive errors (Supplementary Note 5), indicating more accurate and balanced epilepsy detection performance across datasets.

## Robustness analysis

We further examined the influence of several critical factors on model performance, i.e., training set size, class imbalance, and base LLM parameters. To evaluate the effect of training set size, we fine-tuned LLMs on randomly sampled subsets comprising 5% to 80% of the available training data. Performance was then assessed on both in-domain and external datasets. As illustrated in Fig. 5a, even with only 5% of the training data, Qwen-2.5-70B fine-tuned on MIMIC-IV achieved an AUC of 0.806 for in-domain evaluation and 0.924 on the UMN dataset. Performance increased substantially as more data were introduced, followed by a plateau as gains diminished at larger sample sizes. We next examined EpiScreen's sensitivity to class imbalance by adjusting the epilepsy-to-PNES ratio in the training data. Model performance remained relatively stable at an imbalance ratio of 5, but a more severe imbalance led to a gradual reduction in performance (Fig. 5b). Finally, we assessed the effect of model parameter scale. Both in-domain and generalization performance of the fine-tuned Qwen models declined progressively as the parameter size decreased from 70B to 0.5B (Fig. 5c).

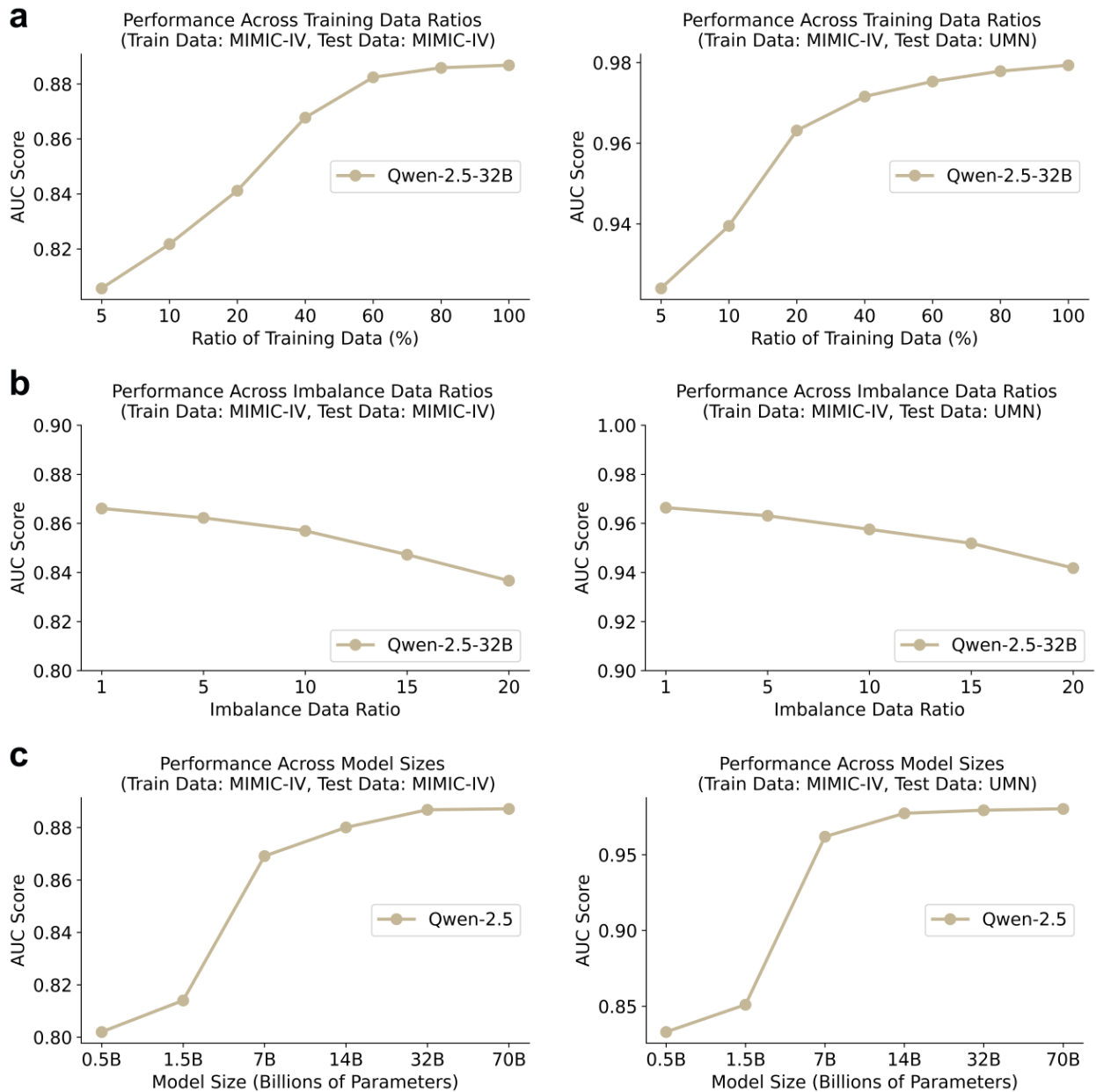

**Fig. 5 | Robustness analysis of the LLM-based epilepsy detection model. a,** Training data ratio analysis evaluating performance as the training set size was progressively reduced from 5% to 100% of the original MIMIC-IV data while preserving class ratios. **b,** Performance under varying epilepsy-to-PNES class imbalance ratios in the training set, with validation and test sets held constant. **c,** Model scaling analysis across Qwen-2.5 models of different parameter sizes. In all experiments, models were fine-tuned on MIMIC-IV clinical notes and evaluated on held-out MIMIC-IV and UMN test sets. These analyses assess robustness to training data availability, class distribution, and model capacity.

# Discussion

In this study, we developed an epilepsy detection approach based solely on routinely documented clinical narratives by fine-tuning LLMs. We highlight the following key observations and insights for further discussion.

First, EpiScreen achieved high diagnostic accuracy in using clinical narratives for identifying epilepsy, significantly outperforming established comparative methods (Table 2). Traditional phenotype-based systems, e.g., FSLS [19], depend on predefined cues and therefore struggle with the probabilistic and overlapping nature of seizure manifestations. In contrast, EpiScreen integrates distributed linguistic signals across the entire note, capturing how clinicians contextualize symptoms, temporal evolution, and comorbidities. This holistic processing reduces overreliance on isolated features and better reflects real-world diagnostic reasoning. Compared to conventional language models, e.g., BERT [30] and ClinicalBERT [32], LLMs benefit from large-scale pretraining that embeds general medical knowledge and linguistic structure, providing a strong foundation for subsequent domain adaptation. Subsequent instruction fine-tuning further aligns representations with epilepsy-specific distinctions, enabling detection of subtle but recurrent cue combinations that are difficult to encode manually. Together, these properties likely contribute to EpiScreen's reduced error rates (Fig. 4) and support its suitability for the complex neuropsychiatric differentiation task.

Second, strong cross-institutional generalization suggests that EpiScreen learned transferable clinical representations rather than site-specific artifacts (Table 3). Clinical documentation varies substantially across institutions in style, terminology, and level of detail, which often limits the portability of conventional models [38,39]. The relative performance stability may stem from the ability of our models to abstract higher-level semantics from heterogeneous phrasing. Instead of anchoring on exact keywords, EpiScreen likely captures relational patterns, such as the linkage between triggers and prodromes (Table 4), that remain conceptually consistent across sites. In addition, exposure to comprehensive pretraining corpora may help normalize linguistic variability and reduce sensitivity to local documentation habits. Fine-tuning on labeled notes further calibrates these representations toward clinically meaningful distinctions [34]. From a translational perspective, generalization performance is critical, as screening tools must function reliably across healthcare systems [40]. These findings support the premise that narrative-based models can form the basis of scalable decision support that generalizes beyond the environment in which they were trained.

Third, explainability analyses indicate that model attributions align with neurologists' diagnostic reasoning. High-attribution sentences frequently reflected seizure semiology, temporal structure, and diagnostic context, all of which are central to neurologists' differentiation between epilepsy and PNES [41,42]. Importantly, EpiScreen did not appear to focus solely on isolated hallmark phrases but instead weighted clusters of related narrative elements (Fig. 2). This pattern suggests that the model is approximating a Gestalt assessment [43] similar to clinician judgment. The prominence of diagnostic testing context and comorbidity information also highlights that differentiation is multifactorial, extending beyond ictal descriptions alone (Table 4 and

Supplementary Note 6). Such alignment between model attributions and clinical logic increases confidence that predictions are grounded in meaningful content rather than spurious correlations [44]. At a broader level, interpretable outputs may facilitate neurologists' trust and enable targeted validation [45]. Explainability, therefore, serves not only as a technical screening tool but also as a bridge for human–AI collaboration in clinical settings [46].

Consistent with this, AI assistance improved clinicians' screening performance, underscoring the complementary strengths of humans and the proposed models (Fig. 3). Rather than replacing expert judgment, EpiScreen could function as a cognitive aid that highlights salient information and reduces oversight (Table 4). In overwhelming clinical workflows, relevant historical details can be dispersed across lengthy notes; automated prioritization of influential sentences may help neurologists focus on the most diagnostically informative segments [47]. Importantly, providing both predictions and explanations allows neurologists to critically appraise model outputs instead of accepting them blindly [48]. Such collaborative frameworks may be especially valuable in early screening contexts, where the goal is risk stratification. This human–AI partnership model may represent a practical pathway for integrating diagnostic models into neurology practice [49].

Robustness analyses further provide insight into when and why EpiScreen is effective (Fig. 5). The finding that reasonable performance emerges even with limited training data likely reflects strong knowledge transfer from pretraining, which supplies general medical and linguistic priors (Fig. 5a). The performance plateau with larger training data suggests diminishing returns once core patterns are learned. Sensitivity to severe class imbalance highlights the importance of representative sampling (Fig. 5b), as skewed distributions can bias probabilistic decision boundaries. The observed relationship between parameter scale and performance is also consistent with scaling laws [50] (Fig. 5c): larger models possess greater capacity to encode nuanced semantic relationships. However, this must be balanced against computational and deployment constraints [51]. Collectively, these factors indicate that both training data quality and model capacity contribute to success. In practice, deploying appropriately scaled LLMs (e.g., ≥32B parameters) may provide a practical balance between predictive accuracy and computational feasibility.

This study has several limitations. First, although we performed cross-institutional validation, only two datasets were included; thus, performance in community or international settings with different documentation styles remains uncertain [52,53]. Second, although we preprocessed clinical narratives to approximate real-world diagnostic scenarios, the study is inherently retrospective. Future work will focus on prospective validation [54] to evaluate the performance of EpiScreen in real-time clinical settings.

In summary, EpiScreen is designed to operate at the specialty-care stage, leveraging routinely documented clinical narratives and reports generated during and after neurology referral to provide earlier diagnostic support before confirmatory video EEG, thereby potentially reducing diagnostic delays by approximately 8–12 weeks. By capturing nuanced contextual signals and generating interpretable outputs, EpiScreen accurately distinguishes epilepsy from PNES, complements clinician judgment, and enhances risk stratification. While not a replacement for definitive testing, EpiScreen may support more informed interim clinical decision-making, such as guiding the urgency and allocation of vEEG monitoring or reducing unnecessary exposure to antiseizure medications in patients likely to have PNES. With further prospective validation and thoughtful integration into workflows, EpiScreen has the potential to become a practical and scalable decision-support tool in neurological care.

# Methods

## Dataset curation

The MIMIC-IV database [31] contains de-identified records of approximately 300,000 individuals treated at Beth Israel Deaconess Medical Center (Boston, MA, USA) between 2008 and 2019, including structured data (e.g., laboratory results and diagnostic codes) and unstructured clinical notes (e.g., radiology reports and discharge summaries). The UMN dataset is a proprietary EHR cohort comprising 32,660 patients treated at the University of Minnesota Medical Center between 2011 and 2024, with comparable structured and narrative clinical documentation.

We excluded patients with coexisting diagnoses of epilepsy and PNES to ensure diagnostic clarity. Case identification followed a multi-step filtering procedure. First, candidate patients were identified using ICD codes from the diagnosis table. Second, discharge summaries were manually reviewed, and only cases in which the primary discharge diagnosis matched one of the target ICD codes were retained. Patients assigned diagnostic codes corresponding to both epilepsy and PNES were excluded to minimize confounding. Detailed ICD codes are provided in Supplementary Table S1–2. Fine-grained codes were consolidated into two categories, i.e., epilepsy and PNES, which served as classification labels. To approximate real-world diagnostic scenarios, clinical notes were further processed to reflect timepoints preceding diagnosis (Supplementary Note 1). The final dataset was randomly partitioned into training, validation, and test sets in a 7:1:2 ratio, stratified by disease category.

## Model training

We developed the epilepsy diagnosis model on open-source LLMs through instruction fine-tuning, where the diagnostic task was formatted as natural language instructions with structured outputs. Clinical notes were converted into instructional demonstrations, each comprising: (1) an input note with patient information, (2) a task instruction, and (3) an output containing the diagnosis. An example of an instruction is shown in Supplementary Note 2. Because standard full-parameter fine-tuning of LLMs is computationally demanding, we employed Quantized Low-Rank Adaptation (QLoRA) [55] for efficient training. In this framework, base model parameters were frozen and quantized to reduce memory requirements, while low-rank trainable adapters were introduced into selected layers to enable task-specific adaptation. This strategy allowed the model to acquire epilepsy-related diagnostic patterns from clinical text while preserving training stability and computational efficiency.

To evaluate the impact of class imbalance, we conducted a controlled imbalance analysis. After an initial fixed split of the dataset, class proportions were modified only in the training set, while validation and test sets were kept constant. Let $t$ denote the initial number of PNES samples in the training set. We resampled the training data to a fixed size of $2t$, and then enforced a predefined epilepsy-to-PNES ratio α by selectively removing samples from one class. The ratio α was set to {1, 5, 10, 15, 20}, where larger values indicate greater class imbalance.

To assess the effect of training data scale, we performed a training data ratio analysis. Using the same fixed data split, we varied only the size of the training set while keeping the validation and test sets unchanged. Specifically, the training set size was adjusted to 5%, 10%, 20%, 40%, 60%, 80%, or 100% of the original training data. At each level, samples were randomly subsampled while preserving the original epilepsy-to-PNES class ratio. This design enabled a systematic evaluation of model performance as a function of available training data.

To investigate the influence of model capacity, we fine-tuned Qwen models [35] of different parameter sizes (0.5B, 1.5B, 7B, 14B, 32B, and 70B) under the same training protocol and evaluated all models on the identical held-out test set. This comparison provided insight into the relationship between model scale and diagnostic performance.

## Implementation details

We trained and evaluated four open-source large language models on real-world clinical notes, including three general-domain LLMs (Llama-3.3-70B [37], Qwen-2.5-32B [35], and DeepSeek-R1-32B [56]) and one medical LLM (Med42-70B [33]). We used the standard cross-entropy loss as a learning objective. Model adaptation was performed using QLoRA [55] within the HuggingFace framework with 4-bit NF4 quantization. Training used a batch size of 1 for 3 epochs, a learning rate of 5e-4 with cosine scheduling, and a warm-up ratio of 0.03. The LoRA configuration included a rank of 16, alpha of 32, and a dropout rate of 0.05, with a maximum sequence length of 4096 tokens. No additional hyperparameter tuning was conducted. To obtain predicted probabilities, we used LLMs' internal confidence scores [57–59]. For each clinical note, the model was prompted to classify the condition as either "Epilepsy" or "PNES". We then extracted the model's confidence scores for both possible outcomes from its final prediction layer. We then applied log-softmax normalization across these two logit values to obtain the final probability. All experiments were run on four NVIDIA A100 GPUs (80 GB VRAM).

For comparison, FSLS [19], BERT [60], and ClinicalBERT [32] were developed as standard supervised baselines on the same dataset. To ensure a fair comparison with long-context models, BERT-based baselines were adapted to process up to 4096 tokens using a sliding-window approach [61], where notes were divided into 512-token segments, independently encoded, and aggregated via mean pooling. We also evaluated LLM zero-shot prompting methods, in which pretrained models were queried with the same diagnostic instructions without parameter updates. Hyperparameters for all comparative methods followed commonly used settings reported in prior work [19,26,30].

## Model explainability

We interpreted LLMs' predictions using a sentence-level Integrated Gradients (IG) approach [36] to determine which parts of a clinical note most influenced the diagnosis. Prior clinical research has identified semiologic and historical features, such as ictal duration or tongue biting, that are statistically associated with epilepsy or PNES [9,24]; however, no single feature is pathognomonic, and clinical judgment typically relies on the broader context in which these features appear [25,62]. Motivated by this, we selected sentences, rather than individual words, as the unit of analysis, because sentences better capture coherent clinical context and reflect how clinicians document and interpret patient histories. We first split each note into sentences and converted every sentence into a numeric representation using the LLM. To estimate importance, we used Integrated Gradients to see how the prediction changes when moving from "empty" information to the actual sentences. In brief, if adding a sentence causes a larger change in the prediction, that sentence is considered more influential. We summarized this influence as a single score per sentence and ranked the sentences to highlight the most clinically relevant evidence.

Because epilepsy diagnosis depends on constellations of clinical features rather than isolated lexical cues [62], we further aggregated sentence-level attributions into clinically defined phenotype categories (Supplementary Note 3). This strategy produces interpretations that better align with neurologic reasoning, reduces lexical noise, and facilitates cohort-level analysis within clinically meaningful domains. Neurologists curated a set of phenotype categories (Supplementary Table S3) based on prior literature describing features that help distinguish epilepsy from PNES, including Demographics & Background, Psychiatric & Psychological Traits, Neurological & Medical Comorbidity, Pre-ictal Symptoms, Ictal Semiology, Post-ictal Features, Longitudinal Temporal Pattern, External Triggers, Injury Consequences, Clinical Evaluation & Diagnostic Findings, and Healthcare Utilization Pattern. We then used an LLM (Llama-3.3-70B) to assign each sentence to its corresponding phenotype category, when applicable. To quantify the contribution of each phenotype category, we first identified the top k sentences with the highest IG scores within each clinical note. For a given phenotype category c, the accumulated IG score was defined as:

$$A_c = \frac{1}{k} \sum_{i=1}^{k} IG_i \cdot 1(i \in c) \qquad (1)$$

where $IG_i$ denotes the IG score of the $i$-th highest-ranked sentence, and $1(i \in c)$ is an indicator function equal to 1 if the sentence belongs to phenotype category $c_i$, and 0 otherwise. Dividing by $k$ normalizes the score to facilitate comparisons across notes. This accumulated IG score reflects the relative contribution of each phenotype category to the model's prediction, thereby highlighting which clinical domains most strongly influence classification and enhancing the interpretability of EpiScreen. In this study, we present these accumulated IG scores for predicted epilepsy cases to demonstrate the model's explainability in identifying epilepsy (Fig. 2 and Supplementary Note 4).

## Expert–AI collaboration

To assess the utility of clinician–AI collaboration [46] in epilepsy detection, we established a framework in which neurologists evaluated clinical notes with support from model outputs. In this setting, the EpiScreen generated a case-level diagnostic prediction, and sentence-level Integrated Gradients attribution scores [36] were displayed to indicate the textual evidence underlying the model's decision. Four clinicians participated, divided by experience level: Group A included two neurology residents, and Group B included two senior neurologists with over ten years of experience. Clinicians had access to internet resources to approximate real-world practice, and predictions were determined by within-group consensus. For evaluation, 300 cases were randomly selected from the test set of MIMIC-IV with stratified sampling.

## Evaluation metrics

We evaluated model performance using the area under the receiver operating characteristic curve (AUC) and classification accuracy [19,30,63]. ROC curves were constructed from predicted probabilities for epilepsy by computing sensitivity and specificity across a range of decision thresholds. AUC quantifies discriminative ability independent of a fixed cutoff and ranges from 0.5 (no discrimination) to 1.0 (perfect discrimination). Accuracy was defined as the proportion of correctly classified cases among all samples. A non-parametric bootstrap with 1000 replicates was used to quantify uncertainty. Each replicate was formed by sampling the test set with replacement to the same size as the original, and the resulting distributions were used to derive mean estimates and 95% confidence intervals for each metric.

# Ethics Statement

This retrospective study was approved by the Institutional Review Board of the University of Minnesota Medical School (Approval No. STUDY00023848) and involved access to identifiable electronic health record data. All procedures were conducted in accordance with the Declaration of Helsinki and relevant institutional guidelines. Access to identifiable data was limited to authorized personnel and handled securely to ensure confidentiality and privacy. The requirement for informed consent was waived due to the retrospective nature of the study and use of existing electronic health records. Additionally, this study used data from the MIMIC-IV database, a publicly available, de-identified critical care database, which does not require additional IRB approval. All analyses complied with the MIMIC-IV data use agreement.

# Data Availability

The MIMIC-IV dataset is available on PhysioNet (https://physionet.org/content/mimiciv/3.1), and access is restricted in accordance with its data use agreements. The UMN dataset was derived from real-world clinical practice and was approved by the relevant Institutional Review Board. Owing to patient privacy and regulatory constraints, the associated clinical notes are not publicly shareable.

# Author Contributions

R.Z., S.Z., and Z.S. conceived the study. S.Z. led the overall study design. S.Z., K.Y., and M.Z. performed the literature review, and S.Z. developed the model. S.Z., Z.S., and R.Z. coordinated and supervised data collection, extraction, and human evaluation. S.Z., K.Y., Z.Z., H.Z., and M.Z. were responsible for data collection, preprocessing, and statistical analysis. S.Z., F.X., and R.Z. designed the experiments. S.Z. prepared the first draft of the manuscript, and R.Z. oversaw the project. All authors contributed to the interpretation of results, critically revised the manuscript, and approved the final version for submission.

# Acknowledgements

This work received financial support from the National Institutes of Health, including the National Center for Complementary and Integrative Health (R01AT009457), the National Institute on Aging (R01AG078154), and the National Cancer Institute (R01CA287413). The content of this article represents the authors' perspectives and does not necessarily reflect the official views of the NIH. Additional support was provided by the Center for Learning Health System Sciences at the University of Minnesota.

# References

1. Kobau, R., Luncheon, C. & Greenlund, K. Active epilepsy prevalence among U.S. adults is 1.1% and differs by educational level-National Health Interview Survey, United States, 2021. *Epilepsy Behav.* **142**, 109180 (2023).
2. GBD Epilepsy Collaborators. Global, regional, and national burden of epilepsy, 1990-2021: a systematic analysis for the Global Burden of Disease Study 2021. *Lancet Public Health* **10**, e203–e227 (2025).
3. Devinsky, O., Gazzola, D. & LaFrance, W. C., Jr. Differentiating between nonepileptic and epileptic seizures. *Nat. Rev. Neurol.* **7**, 210–220 (2011).
4. Liampas, A., Markoula, S., Zis, P. & Reuber, M. Psychogenic non-epileptic seizures (PNES) in the context of concurrent epilepsy – making the right diagnosis. *Acta Epileptol.* **3**, (2021).
5. Durrant, J., Rickards, H. & Cavanna, A. E. Prognosis and outcome predictors in psychogenic nonepileptic seizures. *Epilepsy Res. Treat.* **2011**, 274736 (2011).
6. Widyadharma, I. P. E., Soejitno, A., Samatra, D. P. G. P. & Sinardja, A. M. G. Clinical differentiation of psychogenic non-epileptic seizure: a practical diagnostic approach. *Egypt. J. Neurol. Psychiatr. Neurosurg.* **57**, (2021).
7. Bodde, N. M. G. *et al.* Psychogenic non-epileptic seizures--diagnostic issues: a critical review. *Clin. Neurol. Neurosurg.* **111**, 1–9 (2009).
8. Bompaire, F. *et al.* PNES Epidemiology: What is known, what is new? *Eur. J. Trauma Dissociation* **5**, 100136 (2021).
9. Perez, D. L. & LaFrance, W. C., Jr. Nonepileptic seizures: an updated review. *CNS Spectr.* **21**, 239–246 (2016).
10. Nowacki, T. A. & Jirsch, J. D. Evaluation of the first seizure patient: Key points in the history and physical examination. *Seizure* **49**, 54–63 (2017).
11. Gedzelman, E. R. & LaRoche, S. M. Long-term video EEG monitoring for diagnosis of psychogenic nonepileptic seizures. *Neuropsychiatr. Dis. Treat.* **10**, 1979–1986 (2014).
12. Vander, T. *et al.* Economic aspects of prolonged home video-EEG monitoring: a simulation study. *Cost Eff. Resour. Alloc.* **22**, 59 (2024).
13. Kerr, W. T. *et al.* Factors associated with delay to video-EEG in dissociative seizures. *Seizure* **86**, 155–160 (2021).
14. Klein, H., Pang, T., Slater, J. & Ramsay, R. E. How much time is enough? Establishing an optimal duration of recording for ambulatory video EEG. *Epilepsia Open* **6**, 569–578 (2021).
15. Connolly, B. *et al.* Assessing the similarity of surface linguistic features related to epilepsy across

pediatric hospitals. *J. Am. Med. Inform. Assoc.* **21**, 866–870 (2014).

16. Pevy, N., Christensen, H., Walker, T. & Reuber, M. Differentiating between epileptic and functional/dissociative seizures using semantic content analysis of transcripts of routine clinic consultations. *Epilepsy Behav.* **143**, 109217 (2023).

17. Yu, K. *et al.* AI enabled decision support systems in epilepsy surgery a scoping review. *Research Square* (2026) doi:10.21203/rs.3.rs-8612799/v1.

18. Zhou, S. *et al.* Large language models for disease diagnosis: a scoping review. *NPJ Artif. Intell.* **1**, 9 (2025).

19. Kerr, W. T. *et al.* Supervised machine learning compared to large language models for identifying functional seizures from medical records. *Epilepsia* **66**, 1155–1164 (2025).

20. Cardeña, E., Pick, S. & Litwin, R. Differentiating psychogenic nonepileptic from epileptic seizures: A mixed-methods, content analysis study. *Epilepsy Behav.* **109**, 107121 (2020).

21. Hamid, H. *et al.* Validating a natural language processing tool to exclude psychogenic nonepileptic seizures in electronic medical record-based epilepsy research. *Epilepsy Behav.* **29**, 578–580 (2013).

22. Opp, J. *et al.* The EpiLing-Tool: A new tool to distinguish epileptic seizures from dissociative seizures in the first encounter between physician and patient. *Seizure* **130**, 100–105 (2025).

23. Zhang, L. *et al.* Automated seizure classification using Multimodal Large Language Models. *medRxiv* (2025) doi:10.1101/2025.10.07.25337538.

24. Brigo, F. *et al.* Tongue biting in epileptic seizures and psychogenic events: an evidence-based perspective. *Epilepsy Behav.* **25**, 251–255 (2012).

25. Leibetseder, A., Eisermann, M., LaFrance, W. C., Jr, Nobili, L. & von Oertzen, T. J. How to distinguish seizures from non-epileptic manifestations. *Epileptic Disord.* **22**, 716–738 (2020).

26. Ford, J., Pevy, N., Grunewald, R., Howell, S. & Reuber, M. Can artificial intelligence diagnose seizures based on patients' descriptions? A study of GPT-4. *Epilepsia* **66**, 1959–1974 (2025).

27. Brigo, F., Broggi, S., Leuci, E., Turcato, G. & Zaboli, A. Can ChatGPT 4.0 diagnose epilepsy? A study on artificial intelligence's diagnostic capabilities. *J. Clin. Med.* **14**, 322 (2025).

28. Elger, C. E. & Hoppe, C. Diagnostic challenges in epilepsy: seizure under-reporting and seizure detection. *Lancet Neurol.* **17**, 279–288 (2018).

29. Saleem, M. N. *et al.* Investigation of patient and observer agreement on description of seizures at initial clinical visit. *Ann. Clin. Transl. Neurol.* **6**, 2601–2606 (2019).

30. Loyens, J. *et al.* AI language model applications for early diagnosis of childhood epilepsy based on unstructured first-visit patient narratives: A cohort study. *Epileptic Disord.* **27**, 1263–1274 (2025).

31. Johnson, A. E. W. *et al.* MIMIC-IV, a freely accessible electronic health record dataset. *Sci. Data* **10**, 1 (2023).

32. Alsentzer, E. *et al.* Publicly available clinical BERT embeddings. *arXiv [cs.CL]* (2019).
33. Christophe, C. *et al.* Med42 -- evaluating fine-tuning strategies for medical LLMs: Full-parameter vs. Parameter-efficient approaches. *arXiv [cs.CL]* (2024).
34. Zhou, S. *et al.* Uncertainty-aware large language models for explainable disease diagnosis. *NPJ Digit. Med.* **8**, 690 (2025).
35. Qwen *et al.* Qwen2.5 Technical Report. *arXiv [cs.CL]* (2024).
36. Sundararajan, M., Taly, A. & Yan, Q. Axiomatic attribution for deep networks. *arXiv [cs.LG]* 3319–3328 (2017) doi:10.5555/3305890.3306024.
37. Grattafiori, A. *et al.* The Llama 3 herd of models. *arXiv [cs.AI]* (2024).
38. Yue, X. & Zhou, S. PHICON: Improving generalization of clinical text DE-identification models via data augmentation. in *Proceedings of the 3rd Clinical Natural Language Processing Workshop* (Association for Computational Linguistics, Stroudsburg, PA, USA, 2020). doi:10.18653/v1/2020.clinicalnlp-1.23.
39. Keloth, V. K. *et al.* Social determinants of health extraction from clinical notes across institutions using large language models. *NPJ Digit. Med.* **8**, 287 (2025).
40. Yang, J., Soltan, A. A. S. & Clifton, D. A. Machine learning generalizability across healthcare settings: insights from multi-site COVID-19 screening. *NPJ Digit. Med.* **5**, 69 (2022).
41. Fernandes, M. *et al.* Extracting seizure control metrics from clinic notes of patients with epilepsy: A natural language processing approach. *Epilepsy Res.* **207**, 107451 (2024).
42. Yew, A. N. J., Schraagen, M., Otte, W. M. & van Diessen, E. Transforming epilepsy research: A systematic review on natural language processing applications. *Epilepsia* **64**, 292–305 (2023).
43. Dale, A. P., Marchello, C. & Ebell, M. H. Clinical gestalt to diagnose pneumonia, sinusitis, and pharyngitis: a meta-analysis. *Br. J. Gen. Pract.* **69**, e444–e453 (2019).
44. Hur, S. *et al.* Comparison of SHAP and clinician friendly explanations reveals effects on clinical decision behaviour. *NPJ Digit. Med.* **8**, 578 (2025).
45. Xue, C. *et al.* AI-based differential diagnosis of dementia etiologies on multimodal data. *Nat. Med.* **30**, 2977–2989 (2024).
46. Wang, G. *et al.* Human-large language model collaboration in clinical medicine: a systematic review and meta-analysis. *NPJ Digit. Med.* **9**, (2026).
47. Tam, T. Y. C. *et al.* A framework for human evaluation of large language models in healthcare derived from literature review. *NPJ Digit. Med.* **7**, 258 (2024).
48. Zhou, S. *et al.* Explainable differential diagnosis with dual-inference large language models. *Npj Health Syst* **2**, 12 (2025).
49. Bhatt, A. B. & Bae, J. Collaborative Intelligence to catalyze the digital transformation of healthcare.

*NPJ Digit. Med.* **6**, 177 (2023).

50. Riedemann, L., Labonne, M. & Gilbert, S. The path forward for large language models in medicine is open. *NPJ Digit. Med.* **7**, 339 (2024).

51. Zhan, Z. *et al.* Quantized large language models in biomedical natural language processing: Evaluation and recommendation. *arXiv [cs.CL]* (2025).

52. Li, B. *et al.* The performance of a deep learning system in assisting junior ophthalmologists in diagnosing 13 major fundus diseases: a prospective multi-center clinical trial. *NPJ Digit. Med.* **7**, 8 (2024).

53. Guo, L. L. *et al.* A multi-center study on the adaptability of a shared foundation model for electronic health records. *NPJ Digit. Med.* **7**, 171 (2024).

54. Marchetti, M. A. *et al.* Prospective validation of dermoscopy-based open-source artificial intelligence for melanoma diagnosis (PROVE-AI study). *NPJ Digit. Med.* **6**, 127 (2023).

55. Dettmers, T., Pagnoni, A., Holtzman, A. & Zettlemoyer, L. QLoRA: Efficient Finetuning of Quantized LLMs. *arXiv [cs.LG]* (2023) doi:10.5555/3666122.3666563.

56. Guo, D. *et al.* DeepSeek-R1 incentivizes reasoning in LLMs through reinforcement learning. *Nature* **645**, 633–638 (2025).

57. Gu, B., Desai, R. J., Lin, K. J. & Yang, J. Probabilistic medical predictions of large language models. *NPJ Digit. Med.* **7**, 367 (2024).

58. Bentegeac, R., Le Guellec, B., Kuchcinski, G., Amouyel, P. & Hamroun, A. Token probabilities to mitigate large language models overconfidence in answering medical questions. *J. Med. Internet Res.* **27**, e64348 (2025).

59. Park, J. *et al.* Enhancing EHR-based pancreatic cancer prediction with LLM-derived embeddings. *NPJ Digit. Med.* **8**, 465 (2025).

60. Devlin, J., Chang, M.-W., Lee, K. & Toutanova, K. BERT: Pre-training of deep bidirectional Transformers for language understanding. *arXiv [cs.CL]* (2018).

61. Li, Y., Wehbe, R. M., Ahmad, F. S., Wang, H. & Luo, Y. A comparative study of pretrained language models for long clinical text. *J. Am. Med. Inform. Assoc.* **30**, 340–347 (2023).

62. Oto, M. M. The misdiagnosis of epilepsy: Appraising risks and managing uncertainty. *Seizure* **44**, 143–146 (2017).

63. Brigo, F. *et al.* Artificial intelligence (ChatGPT 4.0) vs. Human expertise for epileptic seizure and epilepsy diagnosis and classification in Adults: An exploratory study. *Epilepsy Behav.* **166**, 110364 (2025).